\pdfoutput=1
\documentclass[11pt,a4paper]{article}
\usepackage{style/eacl2023}
\usepackage{times}
\usepackage{latexsym}
\usepackage{url}
\usepackage{booktabs}
\usepackage{tikz}
\usepackage[utf8]{inputenc}
\usepackage[T1]{fontenc}
\usepackage{anyfontsize}
\usepackage{caption}
\usepackage{subcaption}
\usepackage{xstring}
\usepackage{todonotes}
\usepackage{csquotes}
\usepackage{float}
\usepackage{pifont}
\usepackage{mathtools}
\usepackage{xcolor}
\usepackage{enumitem}
\usepackage{diagbox}
\usepackage{microtype}
\usepackage{inconsolata}
\newcommand{\fireemoji}[0]{\includegraphics[width=.023\textwidth]{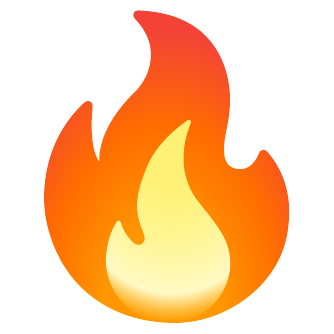}}

\newcommand{\projecturl}{\href{https://www.pedro.ai/multimodal-retrieval-evaluation}{pedro.ai/multimodal-retrieval-evaluation}}
\newcommand{\ir}{\abr{ir}}
\newcommand{\msrvtt}{\abr{msr}-\abr{vtt}}

\newcommand{\msvd}{\abr{msvd}}
\newcommand{\ssb}{\abr{ssb}}
\newcommand{\anet}{ActivityNet}

\newcommand{\firesize}{683K}
\newcommand{\clip}{\abr{c4c}}
\newcommand{\clipl}{\abr{clip4clip}}
\newcommand{\teachtextl}{TeachText}
\newcommand{\teachtext}{\abr{tt}}
\newcommand{\mmr}{\abr{tvr}}
\newcommand{\fire}{\abr{fire}}
\newcommand{\ttv}{text-to-video}

\newcommand{\nmsrvttlabels}{0}
\newcommand{\nmsvdlabels}{0}
\newcommand{\nmsrvttpairs}{0}
\newcommand{\nmsvdpairs}{0}
\newcommand{\nmsrvttdisagree}{0}
\newcommand{\pmsrvttdisagree}{0}
\newcommand{\nmsvddisagree}{0}
\newcommand{\pmsvddisagree}{0}
\newcommand{\nmsrvtttotal}{0}
\newcommand{\nmsvdtotal}{0}
\newcommand{\nmsrvttpositive}{0}
\newcommand{\pmsrvttpositive}{0}
\newcommand{\nmsrvttnegative}{0}
\newcommand{\pmsrvttnegative}{0}
\newcommand{\nmsvdpositive}{0}
\newcommand{\pmsvdpositive}{0}
\newcommand{\nmsvdnegative}{0}
\newcommand{\pmsvdnegative}{0}
\newcommand{\pmsrvttpairs}{0}
\newcommand{\pmsvdpairs}{0}

\newcommand{\tablewhitespace}{\addlinespace[0.3em]}
\definecolor{altair_blue}{HTML}{1f77b4}
\definecolor{altair_orange}{HTML}{ff7f0e}
\definecolor{altair_green}{HTML}{2ca02c}
\definecolor{altair_red}{HTML}{d62728}
\definecolor{altair_cyan}{HTML}{17becf}

\IfFileExists{auto_fig/dataset-stats.tex}{

\renewcommand{\nmsrvttlabels}{24,507}
\renewcommand{\nmsvdlabels}{659,126}
\renewcommand{\nmsrvttpairs}{24,167}
\renewcommand{\pmsrvttpairs}{99.9}
\renewcommand{\nmsvdpairs}{553,832}
\renewcommand{\pmsvdpairs}{99.7}
\renewcommand{\nmsrvttdisagree}{16}
\renewcommand{\pmsrvttdisagree}{0.0662}
\renewcommand{\nmsvddisagree}{1,559}
\renewcommand{\pmsvddisagree}{0.281}
\renewcommand{\nmsrvtttotal}{24,183}
\renewcommand{\nmsvdtotal}{555,391}
\renewcommand{\nmsrvttpositive}{2,855}
\renewcommand{\pmsrvttpositive}{11.8}
\renewcommand{\nmsrvttnegative}{21,312}
\renewcommand{\pmsrvttnegative}{88.2}
\renewcommand{\nmsvdpositive}{39,909}
\renewcommand{\pmsvdpositive}{7.21}
\renewcommand{\nmsvdnegative}{513,923}
\renewcommand{\pmsvdnegative}{92.8}

}{}

\newif\ifcomment
\commenttrue
\usepackage[a-1b]{pdfx}

\usepackage{framed}
\usepackage{mdwlist}
\usepackage{siunitx}
\usepackage{latexsym}
\usepackage{colortbl}
\usepackage{xcolor}
\usepackage{nicefrac}
\usepackage{booktabs}
\usepackage{fnpct}
\usepackage{amsfonts}
\usepackage[T1]{fontenc}
\usepackage{bold-extra}
\usepackage{amsmath}
\usepackage{amssymb}
\usepackage{bm}
\usepackage{graphicx}
\usepackage{mathtools}
\usepackage{microtype}
\usepackage{multirow}
\usepackage{multicol}
\usepackage{xpatch}
\usepackage{latexsym,comment}
\usepackage[normalem]{ulem}

\newcommand*{\missingreference}{{\Huge \colorbox{red}{?reference?}}}
\newcommand*{\missingcitation}{{\Huge \colorbox{red}{?citation?}}}

\makeatletter
\xpatchcmd{\@setref}{\bfseries}{\missingreference}{}{}
\def\@citex[#1]#2{\leavevmode
    \let\@citea\@empty
    \@cite{\@for\@citeb:=#2\do
        {\@citea\def\@citea{,\penalty\@m\ }%
            \edef\@citeb{\expandafter\@firstofone\@citeb\@empty}%
            \if@filesw\immediate\write\@auxout{\string\citation{\@citeb}}\fi
            \@ifundefined{b@\@citeb}{\hbox{\reset@font\missingcitation}%
                \G@refundefinedtrue
                \@latex@warning
                {Citation `\@citeb' on page \thepage \space undefined}}%
            {\@cite@ofmt{\csname b@\@citeb\endcsname}}}}{#1}}
\makeatother

\newcommand{\gem}[1]{\mbox{\textsc{gem}}}
\newcommand{\abr}[1]{\textsc{#1}}

\newcommand{\emaillink}[1]{{\small \href{mailto://#1}{\texttt{#1}}}}

\newcommand{\hidetext}[1]{}
\newcommand{\ignore}[1]{}

\ifcomment
    \newcommand{\pinaforecomment}[3]{\colorbox{#1}{\parbox{.8\linewidth}{#2: #3}}}
    \newcommand{\prtodo}[1]{\todo[backgroundcolor=lightblue,linecolor=lightblue,author=Pedro]{#1}}
    \newcommand{\prtodoi}[1]{\todo[inline,backgroundcolor=lightblue,linecolor=lightblue,author=Pedro]{#1}}

    \newcommand{\smtodo}[1]{\todo[backgroundcolor=orange,linecolor=orange,author=Shane]{#1}}
    \newcommand{\smtodoi}[1]{\todo[inline,backgroundcolor=orange,linecolor=orange,author=Shane]{#1}}

\else
    \newcommand{\pinaforecomment}[3]{}
    \newcommand{\prtodo}[1]{}
    \newcommand{\prtodoi}[1]{}

    \newcommand{\smtodo}[1]{}
    \newcommand{\smtodoi}[1]{}    
\fi

\newcommand{\smallurl}[1]{ \begin{tiny}\url{#1}\end{tiny}}

\definecolor{lightblue}{HTML}{3cc7ea}
\definecolor{CUgold}{HTML}{CFB87C}
\definecolor{grey}{rgb}{0.95,0.95,0.95}
\definecolor{ceil}{rgb}{0.57, 0.63, 0.81}
\definecolor{UMDred}{HTML}{ed1c24}
\definecolor{UMDyellow}{HTML}{ffc20e}

\newcommand{\nlp}[0]{\abr{nlp}}

\graphicspath{{figures/}{commit_auto_fig/}{auto_fig/}}

\title{Fighting FIRe with FIRE:\\Assessing the Validity of Text-to-Video Retrieval Benchmarks}

\author{
{\bf Pedro Rodriguez}{\normalfont ,}\thanks{\ \ Correspondence to \emaillink{me@pedro.ai}}\\
{\bf Mahmoud Azab}{\normalfont ,}
{\bf Becka Silvert}{\normalfont ,}
{\bf Renato Sanchez}{\normalfont ,}\\
{\bf Linzy Labson}{\normalfont ,}
{\bf Hardik Shah}{\normalfont ,}
{\bf Seungwhan Moon} \\
\tablewhitespace
Meta AI
}

\date{}

\begin{document}
\maketitle
\setlength{\abovedisplayskip}{3pt}
\setlength{\belowdisplayskip}{3pt}

\begin{abstract}
    Searching troves of videos with textual descriptions is a core multimodal retrieval task.
Owing to the lack of a purpose-built dataset for text-to-video retrieval, video captioning datasets have been re-purposed to evaluate models by (1) treating captions as positive matches to their respective videos and (2) assuming all other videos to be negatives.
However, this methodology leads to a fundamental flaw during evaluation: since captions are marked as relevant \textit{only} to their original video, many alternate videos \textit{also} match the caption, which introduces false-negative caption-video pairs.
We show that when these false negatives are corrected, a recent state-of-the-art model gains 25\% recall points---a difference that threatens the validity of the benchmark itself.
To diagnose and mitigate this issue, we annotate and release \firesize{} additional caption-video pairs.
Using these, we recompute effectiveness scores for three models on two standard benchmarks (\msrvtt{} and \msvd{}).
We find that (1) the recomputed metrics are up to 25\% recall points higher for the best models, (2) these benchmarks are nearing saturation for Recall@10, (3) caption length (generality) is related to the number of positives, and (4) annotation costs can be mitigated through sampling.
We recommend retiring these benchmarks in their current form, and we make recommendations for future text-to-video retrieval benchmarks.

\end{abstract}

\section{Introduction}
\label{sec:intro}

% Deal with the author/email star technically being footnote 1
\setcounter{footnote}{0}
Text-to-video retrieval (\mmr{}) is a challenging multimodal retrieval task~\citep{hu2011survey} with practical applications ranging from web search to organizing media collections~\citep{lew2006multi}.
To measure \mmr{} model improvement---despite a dearth of purpose-built \mmr{} benchmarks---researchers created benchmarks by re-purposing video captioning datasets such as \msrvtt{}~\citep{xu2016msrvtt}, \msvd{}~\citep{chen2011msvd}, and \anet{}~\citep{heilbron2015anet,krishna2017anet}.
Early work established an evaluation paradigm that treated captions as search queries over the collection of captioned videos~\citep{zhang2018crossmodal,yu2018fusion,gabeur2020mmt}; each caption and their corresponding video are positives (relevant) during retrieval, and all other caption-video pairs are negatives (irrelevant).

\begin{figure}
    \centering
    \includegraphics[width=\linewidth]{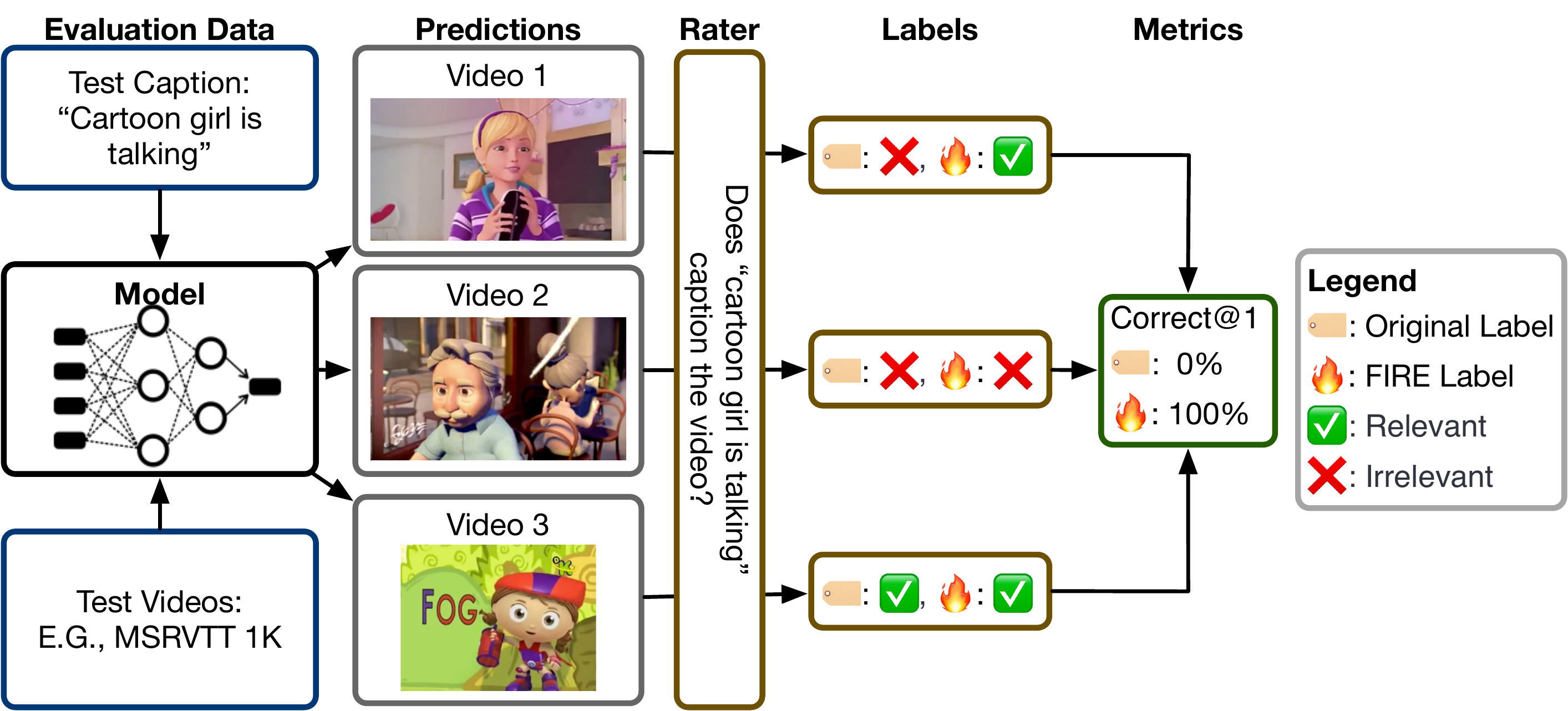}
    \caption{
        \msrvtt{} and \msvd{} have one positive video per caption (each video's caption).
        Captions often match multiple videos, leading to false negatives.
        When models rank false negatives highly, model quality is understated (full example in Appendix Figure~\ref{fig:streamlit-compare}).
        This leads to evaluations where reported metrics do not reflect their true value and are therefore not internally valid~(\S\ref{sec:internal}).
    }
    \label{fig:problem}
\end{figure}

However, even a cursory inspection of videos and captions reveals many additional positive caption-video pairs~(\S\ref{sec:state}).
In current benchmarks, \textit{true positives} that are not the video's original caption are falsely assumed to be \textit{negatives}.
\citet{wray2021similarity} first identified this fundamental, false-negative problem in \mmr{} evaluation; our work builds on this by quantifying the absolute metric differences that false negatives induce (see discussion in \S\ref{sec:rel}).
Accurate absolute metrics are crucial in industrial settings where deployment criteria are often defined by minimum quality targets.
% Unfortunately, frequent queries \textit{also} tend to be the most general captions~\citep{petersen2016power}, leading to a higher false negative rate.
These \textbf{F}alse \textbf{I}mplicit \textbf{Re}levance labels introduce measurement error---e.g., \clipl{}'s~\citep{luo2021clip4clip} Recall@1 is underestimated by 25\% points~(\S\ref{sec:valid}).
We estimate measurement error by annotating \firesize{} additional caption-video pairs, which we call the \fire{}~\fireemoji~dataset~(\S\ref{sec:improving}).\footnote{
    Data and Code: \projecturl{}.
}

A core measurement principle is that operationalized metrics should strongly correlate to the quantity they intend to measure~\citep{mathison2004encyclopedia,liao2021we}.
For example, Recall@K operationalizes the intent to measure retrieval quality.
Label errors are a common way that measurements are invalidated~\citep{bowman2021fix,northcutt2021pervasive}.
Our work shows that since \mmr{} metrics are computed with false negative label errors, Recall@K does not accurately reflect retrieval quality, which negates the measurement's validity.
In the remainder of this paper, we posit rationales of why models gain different score boosts~(\S\ref{sec:boosts}) and estimate how useful the \fire{} dataset is for evaluating future models~(\S\ref{sec:pool-gen} and \S\ref{sec:sampling}).

To conclude, we review the implications of our findings.
Looking to the past, retrieval effectiveness has been understated for some models, which gives an overly pessimistic view of recent advances~\citep{bowman2021danger}.
Critically, our results also suggest that the \msrvtt{} benchmark is nearing saturation and should be retired soon in favor of a purpose-made benchmark.
Looking outward, we identify structurally similar benchmarks---such as photo retrieval---that likely also have the same \textbf{F}alse \textbf{I}mplicit \textbf{Re}levance problem.
A successful benchmark should avoid the pitfalls we identify in this paper, be faithful to the real-world user task it targets~\citep{rowe2005grand,devries2020eco}, improve reproducibility, and evolve~(\S\ref{sec:conc}).

\section{Text-to-Video Retrieval Evaluation}
\label{sec:state}

This section reviews current \mmr{} evaluation practices using two concepts: \textit{internal validity}~\citep[\S\ref{sec:internal}]{campbell1957validity} and \textit{construct validity}~\citep[\S\ref{sec:construct}]{sutcliffe1992pragmatics}.
\textit{Internal validity} refers to whether an evaluation reliably establishes a cause-effect relationship between the measured dependent variable and the independent variable to be estimated~\citep{brewer2014validity,liao2021we}.
In \mmr{} evaluations, false negatives confound model quality and label errors (i.e., is the model wrong or is the label wrong?) which makes \textit{reliably} establishing cause (model quality) and effect (retrieval score) difficult.
\textit{Construct validity} ``pertains to the degree to which the measure of a construct sufficiently measures the intended concept''~\citep{oleary1998validity}---in \mmr{} evaluations, an important intended concept is real-world search quality.
\textit{Construct validity} asks: can we expect that measuring retrieval quality with the benchmarks at hand generalizes to real-world search quality?
This section argues that \mmr{} evaluations are not internally valid or construct valid.

\subsection{Model Evaluation}
\label{sec:setup}

Multimodal retrieval evaluations typically focus on two tasks: text-to-video and video-to-text retrieval.
The first task's goal is---given a text query---to retrieve videos that match; the second task's goal is---given a video---to retrieve the matching queries.
The applications of text-to-video search are straightforward: it is useful for searching the web and personal media.\footnote{
    The applications of video-to-text retrieval---that are not simply captioning---are not clear to us.
}
Since the applications of \mmr{} are clear, and the false-negative problem is present in both tasks, here we focus on \mmr{}.

\textbf{The MSR-VTT and MSVD Datasets:}
It is standard for \mmr{} evaluations~\citep{zhang2018crossmodal,yu2018fusion,gabeur2020mmt} to report on \msrvtt{} and \msvd{}, so in the interest of comparability, we use these benchmarks too.
Although these datasets were originally meant for evaluating video captioning models, they have been repurposed for \mmr{}~\citep{zhang2018crossmodal,gabeur2020mmt}.
In this paper, we focus our investigation on \msrvtt{} and \msvd{} since they are the most prevalent in prior work.
\msrvtt{} consists of 10K videos, 1K of which are in the test split.
Each video has twenty captions, but for evaluation, only one (arbitrarily chosen) caption is used.
\msvd{} contains 1,970 videos, 960 of which are in the test split.
Videos have about forty captions; unlike with \msrvtt{}, retrieval quality for each caption is evaluated.

Fundamentally, both \msrvtt{} and \msvd{} are video captioning datasets---not retrieval datasets.
\msvd{} addressed the lack of standard benchmarks for paraphrasing~\citep{chen2011msvd}.
In the original task, annotators selected short clips from YouTube, watched the clip, and wrote a sentence describing its contents. The process was repeated for each video, with each sentence being written by a new annotator.
This conditional independence---given the video---resulted in a diverse set of captions.
\msrvtt{} captions were collected similarly: independent annotators captioned the same video.
Videos were sourced from the output of a commercial video search engine~\citep{xu2016msrvtt}.
In both datasets, video captions are used as search queries and labeled relevant to the original video.

\textbf{Metrics:}
Previous \mmr{} work~\citep{zhang2018crossmodal,yu2018fusion,gabeur2020mmt,luo2020univl,zhu2020actbert,li2020hero,xu2021vlm,park2022ncl} reports Recall@K (R@K)\footnote{Typical K values include 1, 5, 10, and 50.} and sometimes supplemental metrics such as median or mean rank of the first correct result.
However, R@K in \mmr{} work differs from the textbook information retrieval definition~\citep[p. 155]{manning2008ir} where
\begin{equation}
    \text{R@K}=\frac{\text{\# retrieved positives in top K}}{\text{\# total positives in collection}}.
\end{equation}
In \mmr{} work, query retrieval results are scored one if a relevant video is in the top K and zero otherwise.
The traditional definition of Recall@K \textit{only} reduces to this when there is \textit{exactly} one positive in the collection but is not comparable when there are multiple positives per caption---as in this case.

With the difference now salient, we avoid confusion by defining a new quantity
Correct@K~(C@K) which is 1 if at least one positive is in the top K and 0 otherwise.
Correct@K naturally reduces to Recall@K---as defined in prior work---when there is exactly one positive, but handles the additional positives in our work.
We recommend reporting Correct@K as well as mean average precision~\citep[\abr{map}]{su2015ap,mitra2018an}, a metric widely used in Information Retrieval.

The drawback of Correct@K---shared by median (or mean) rank to first positive---is that it does not directly factor in rank order when there are multiple positives in retrieved results, only coarsely factoring in rank via K value.
\abr{map}~\citep[p. 19]{mitra2018an} is calculated by taking the mean of
\begin{equation}
    \text{AvgPrec}_q=\frac{\sum_{\langle i,v\rangle\in R_q} \text{Prec}_{q,i}\times \text{rel}_q(v)}{\sum_{v\in V}\text{rel}_q(v)}
\end{equation}
for each test query $q$ where $i$ is a video's position in the ranked list $R_q$ of videos, $v$ is a video in collection $V$, and $\text{rel}_q(v)$ denotes whether query $q$ is relevant to video $v$.
Intuitively, this translates to calculating the mean of Precision@K for every K where a positive occurs in ranked predictions $R_q$.
In all experiments, we report Correct@K and \abr{map}.

\subsection{Questioning the Validity of Evaluations}
\label{sec:valid}

In this section, we experimentally argue that current \mmr{} evaluations are not \textit{internally valid}.
Then we argue that they are not \textit{construct valid} by considering actual use-cases for video search.

%If evaluation metrics are construct valid, then they should generalize beyond the scope of a given dataset.

\subsubsection{Internal Validity}
\label{sec:internal}

If an evaluation metric is internally valid~\citep{liao2021we}, then model effectiveness (cause) should be \textit{accurately and reliably} reflected in metrics (effect)~\citep{brewer2014validity}.
%Internal validity ``refers to the truth value that can be assigned to the conclusion that a cause-effect relationship between an independent variable and a dependent variable has been established''~\citep{campbell1957validity,brewer2014validity}.\prtodo{check quote/cite or remove}
A central hypothesis of this paper is that the prevalence of false negatives invalidates the cause-effect relationship between measured model effectiveness and actual effectiveness--i.e., that correcting false negatives will significantly change metrics.\footnote{
    We do not see rank changes in our three models, but score differences suggest that ranks may change with more models.
}

To test this hypothesis, we build the \fire{} dataset, which \textbf{F}ixes \textbf{I}mplicit \textbf{R}elevance \textbf{E}rrors.
We detail the dataset later (\S\ref{sec:improving}), but in short, we take strong retrieval models from the past few years and annotate their top ten predictions on both \msrvtt{} and \msvd{}.
This process---called system pooling---has been used for decades in information retrieval~\citep{spark1975report} and, by construction, eliminates implicit false negatives.\footnote{
    By implicit, we mean false negative from the lack of labeling and presuming non-positives are (implicitly) negative.
    There may still be false negatives arising from human error during annotation.
}
For \msrvtt{}, we collect annotations from \teachtextl{}~\citep{croitoru2021teach}, Support-Set Bottlenecks~\citep[\ssb{}]{patrick2021ssb}, and \clipl{}~\citep{luo2021clip4clip} models; for \msvd{}, we collect annotations from \teachtextl{} and \clipl{} models.\footnote{
    We prioritize models that are (1) publicly available and (2) have sufficient documentation to reproduce.
}\footnote{
    Annotating \msrvtt{} predictions translates to 1,000 * 10 = 10K annotations since only one caption per video is used.
    This is easy compared to \msvd{} annotation, which uses tens of captions per video.
}
Next, we compute model scores using the original positives and compare them to scores calculated with \emph{both} the original positives \emph{and} the new positives in \fire{}.

\begin{table*}[t]
    \small
    \centering
    \begin{tabular}{lllll}
\toprule
  Dataset & Metric &             TeachText &                 \ssb{} &             \clipl{} \\
\midrule
\msrvtt{} &    C@1 & 24.1 (23.3 + 0.800)\% &  27.3 (26.8 + 0.500)\% & 67.4 (42.4 + 25.0)\% \\
\msrvtt{} &    C@5 &  53.2 (50.9 + 2.30)\% &   55.9 (54.5 + 1.40)\% & 90.7 (70.4 + 20.3)\% \\
\msrvtt{} &   C@10 &  67.0 (64.8 + 2.20)\% &   68.9 (66.3 + 2.60)\% & 95.7 (80.2 + 15.5)\% \\
\msrvtt{} &     AP & 36.1 (35.8 + 0.296)\% & 39.3 (39.2 + 0.0374)\% & 69.5 (54.9 + 14.7)\% \\
\midrule
\msvd{} &  C@1 & 34.7 (19.6 + 15.2)\% & Not Annotated & 65.3 (46.6 + 18.8)\% \\
\msvd{} &  C@5 & 64.7 (48.9 + 15.8)\% & Not Annotated & 89.6 (76.8 + 12.8)\% \\
\msvd{} & C@10 & 76.1 (63.9 + 12.2)\% & Not Annotated & 94.0 (85.4 + 8.61)\% \\
\msvd{} &   AP & 44.3 (33.1 + 11.2)\% & Not Annotated & 71.3 (59.7 + 11.6)\% \\
\bottomrule
\end{tabular}
    \caption{
        The table shows the impact of \fire{} annotations on \msrvtt{} and \msvd{} text-to-video retrieval metrics.
        ``A (B + C)'' has metrics computed with \fire{} positives (A), only original positives (B), and the delta (C).
        The deltas emphasize the deleterious effects of false negatives: \clipl{}'s C@1 on \msrvtt{} is understated by 25\% points.
    }
    \label{tab:change}
\end{table*}

Table~\ref{tab:change} clearly demonstrates that \fire{} annotations reveal large metric differences in both \msrvtt{} and \msvd{}.
For example, the C@1 score of \clipl{} is understated by 25\% points, and its C@10 score arguably saturates the benchmark at 95.7\%.
Even ``small'' differences such as those for \teachtextl{} and \ssb{} are on par with the differences used to claim state-of-the-art results.
False negatives directly cause high measurement error, which invalidates the internal validity of the benchmark.

\subsubsection{Construct Validity}
\label{sec:construct}

In addition to problems with internal validity, we posit that \mmr{} evaluations are also not \textit{construct valid}~\citep{cronbach1955construct,oleary1998validity}.
Construct validity is related to ``how closely our evaluations hit the mark in appropriately characterizing the actual anticipated behaviour of the system in the real world or progress on stated motivations and goals for the field''~\citep{raji2021bench}.
What is the real-world use of \ttv{} retrieval (or alternatively, the field's motivations)?
Consider the most straightforward answer: that such systems will be used by users to search through video collections, whether on the web or in personal collections.
First, search queries issued by real users are very likely not similar to captions written by crowd annotators; this is easily observed by inspecting captions in Table~\ref{tbl:caption_examples_msvd_short} and Appendix Table~\ref{tbl:caption_examples_msrvtt}.
Second, the video distribution is unlikely to reflect real use-cases as they were selected by annotators or are search results from seed queries.
Due to these problems, it seems unlikely that the evaluations are construct valid, and future benchmarks should improve this by building evaluations that match the intended use of models---i.e., be ecologically valid~\citep{devries2020eco}.

\section{FIRE Dataset Collection and Validation}
\label{sec:improving}

%Next, we describe \fire{}'s data collection process, summary statistics, and validate its quality.
Next, we describe and analyze the \fire{} dataset.

\subsection{Annotation Task and Dataset Collection}
\label{sec:dataset}

In the \fire{} annotation task, annotators mark whether the displayed caption is relevant to the displayed video.
Implicitly, the caption's video is relevant to it, but how do we judge whether another arbitrary video is relevant?
In other words, how should annotators mark whether a caption is relevant to a video?
In both datasets~(\S\ref{sec:setup}), the caption must be completely consistent with the video; otherwise, it would not be an accurate caption.
Therefore, we enforce the same condition in our task to preserve the original relevance semantics.\footnote{
    Requiring complete matches makes the annotation task easier by eliminating ambiguous partial match cases.
}

Annotators are instructed to mark a caption as relevant to a video only if every element mentioned in the query could be reasonably considered present.
Elements included persons, objects, locations, and activities, as well as quantifiers, qualifiers, and adjectives.
Raters are given some leeway to use interpretation and inference but instructed to err in favor of not relevant if the caption is ambiguous or vague.
For example, for the caption ``a boy playing the violin,'' the video must show a boy who is playing the violin, not a video of only violins or a video with only a boy.
Screenshots of the annotation interfaces and details of sensitive category handling are in Appendix~\ref{apx:annotation}.
Complete annotation guidelines are included in supplemental materials.

To select caption-video pairs to annotate, we obtain the top ten \msrvtt{} and \msvd{} test set predictions from three models: \clipl{}~\citep{luo2021clip4clip}, \abr{ssb}~\citep{patrick2021ssb}, and \teachtextl~\citep{croitoru2021teach}.
For \teachtextl{}, we use model checkpoints available on their webpage.
For \clipl{} and \abr{ssb}, checkpoints are not available, so we train new models and verify that retrieval quality is on par with the literature~(see Table~\ref{tab:change}).

\begin{table}[t]
    \centering
    \small
    \begin{tabular}{l r r r}
        \toprule
        \textbf{Dataset} & \textbf{\# Pairs} & \textbf{Percent} &\textbf{\# Labels} \\%& \textbf{Agreement} & \textbf{Alpha}\\
        \msrvtt{} & \nmsrvtttotal{} & 100\% & \nmsrvttlabels{} \\%&  \agreeratemsrvtt{} &\kripmsrvtt{}\\
        $\llcorner$ Agreement & \nmsrvttpairs{} & \pmsrvttpairs{}\% & - \\%& \abr{na} & \abr{na}\\
        \rowcolor{altair_green!30} \hspace{8pt}$\llcorner$ Relevant & \nmsrvttpositive{} & \pmsrvttpositive{}\% & - \\%& \abr{na} & \abr{na}\\
        \rowcolor{altair_green!30} \hspace{8pt}$\llcorner$ Irrelevant & \nmsrvttnegative{} & \pmsrvttnegative{}\% & -\\% & \abr{na} & \abr{na}\\
        $\llcorner$ Disagreement & \nmsrvttdisagree{} & \pmsrvttdisagree{}\% & - \\%& \abr{na} & \abr{na}\\
        \midrule
        \msvd{} & \nmsvdtotal{} & 100\% & \nmsvdlabels{} \\%&  \agreeratemsvd{} & \kripmsvd{}\\
        $\llcorner$ Agreement & \nmsvdpairs{} & \pmsvdpairs{}\% & - \\%& \abr{na} & \abr{na}\\
        \rowcolor{altair_green!30} \hspace{8pt}$\llcorner$ Relevant & \nmsvdpositive{} & \pmsvdpositive{}\% & - \\%\\&\abr{na} & \abr{na}\\
        \rowcolor{altair_green!30} \hspace{8pt}$\llcorner$ Irrelevant & \nmsvdnegative{} & \pmsvdnegative{}\% & -\\%&\abr{na} & \abr{na}\\
        $\llcorner$ Disagreement & \nmsvddisagree{} & \pmsvddisagree{}\%& - \\%& \abr{na} & \abr{na}\\
        \bottomrule
    \end{tabular}
    \caption{
        The \fire{} dataset is composed of labels for \msrvtt{} and \msvd{} text-video pairs.
        The positive-to-negative ratio is skewed, reflecting that queries do not match most videos.
        We multiply annotate a subset to compute annotator agreement rates and Krippendorff's $\alpha$.
        Agreement on \msrvtt{} was .931 with $\alpha=.691$ and on \msvd{} was .958 with $\alpha=.798$.
        %We see excellent agreement rates with a reasonable Krippendorff's $\alpha$.
        Appendix~\ref{apx:quality} disaggregates agreement rates which are consistent.
    }
    \label{tbl:stats}
\end{table}

Table~\ref{tbl:stats} summarizes the resulting \fire{} dataset.
During data collection, \firesize{} labels were collected across a set of 579K unique caption-video pairs.
Some duplication was intentional: we obtained a second label for 10\% of annotations, and if the labels disagreed, we collected a third label to resolve the disagreement.
Elsewhere, duplication was unintentional: for \msvd{} we did not deduplicate caption-video pairs between two models, so where the predictions overlapped, we obtained additional labels.
Fortunately, this provided an unexpected opportunity to further validate dataset quality.

\subsection{Dataset Quality Validation}
\label{sec:quality}

Before, throughout, and after the collection, we took steps to collect high-quality data and validate its quality.
The annotation task was completed by a team of one hundred raters specifically trained to review caption-video pairs and assess relevance.
These annotators completed a 1,000 job training queue, which was reviewed by data quality leads and this paper's authors.
This allowed annotators to learn to annotate according to our guidelines, request clarification to the guidelines, and request tooling improvements.
Annotators could also escalate tasks for being too ambiguous or confusing, which occurred less than $0.0001\%$ of the time.

After the dataset was collected, we computed three measures of quality in Table~\ref{tbl:stats}: (1) the rate that judgments resolved to a label~(Percent), (2) the degree to which examples with multiples labels agreed~(Agreement), and (3) the Krippendorff alpha score amongst examples with multiple labels~\citep{kripp2004}.
Caption-video pairs resolved to a label $99.9\%$ of the time in \msrvtt{} and $99.6\%$ of the time in \msvd{}.
Agreement in both datasets exceeded $90\%$, and the Krippendorff score suggests reasonable agreement as well.
Based on this analysis, we see no evidence of data quality issues.
The next section digs deeper into \fire{} and suggests explanations for the observed phenomena.

% From the final data set, we did a second rating on 10\% of the data, if the results did not agree, it was sent to a third rater for resolution. The overall accuracy of the data undergoing multi review was 97.26\%.
% In addition, raters were supported by trained quality assurance leaders and had the opportunity to self-escalate if the job was ambiguous or confusing. This occurred at a negligible rate (<0.0001\%.) 

\section{Analysis Experiments}
\label{sec:exp}

The difference \fire{} makes on metrics (Table~\ref{tab:change}) is striking, which begs the question: \textit{why} are there such \textit{large} differences?
We suggest explanations for these differences (\S\ref{sec:boosts}) while investigating how these metrics vary under commonplace evaluation settings such as new model development~(\S\ref{sec:pool-gen}).
% We then test whether automatic text matching schemes would be effective as a stopgap in the place of human annotation~(\S\ref{sec:matching}).

\subsection{Why Are Score Boosts Not Uniform?}
\label{sec:boosts}

\begin{table}[t]
    \small
    \centering
    \begin{tabular}{llll}
\toprule
  Dataset &                  Models & Overlap &    RBO \\
\midrule
\msrvtt{} &       \clip{} \& \ssb{} &  0.0638 & 0.0568 \\
\msrvtt{} & \clip{} \& \teachtext{} &  0.0610 & 0.0509 \\
\msrvtt{} &  \teachtext{} \& \ssb{} &   0.440 &  0.231 \\
  \msvd{} & \clip{} \& \teachtext{} &   0.411 &  0.211 \\
\bottomrule
\end{tabular}

    \caption{
        Annotated predictions of one model boost the score of another model when predictions overlap.
        In \msrvtt{}, there is little overlap between \clipl{} and other models; there is far more overlap in \msvd{}.
    }
    \label{tab:overlap}
\end{table}
\begin{table}
    \small
    \centering
    \begin{tabular}{llrrr}
\toprule
        Model & Data &   C@1 &   C@5 &  C@10 \\
\midrule
     \clipl{} &  All & 0.674 & 0.907 & 0.957 \\
     \clipl{} &  New & 0.430 & 0.713 & 0.812 \\
\midrule
\teachtextl{} &  All & 0.241 & 0.532 & 0.670 \\
\teachtextl{} &  New & 0.239 & 0.527 & 0.663 \\
\midrule
       \ssb{} &  All & 0.273 & 0.559 & 0.689 \\
       \ssb{} &  New & 0.271 & 0.553 & 0.679 \\
\bottomrule
\end{tabular}

    \caption{
        We compare C@K of a \msrvtt{} model: (1) with all annotations (All) and (2) without the model's annotated predictions to emulate model development (New).
        \clipl{} exhibits large differences.
    }
    \label{tab:ablation}
\end{table}

\fire{}-based metrics are interesting for at least two reasons: (1) the magnitude of difference and (2) the non-uniformity of boosts.
Specifically, \clipl{} has a larger boost than \teachtextl{} and \ssb{} on \msrvtt{}.
First, we investigate the degree of prediction overlap between models.
When predictions overlap, the models share the boost.
Likewise, when they do not overlap, there is an opportunity for differing boosts.
Table~\ref{tab:overlap} shows this: on \msrvtt{}, \clipl{} and the other two models have little overlap; in contrast, \teachtextl{} and \ssb{} have substantial overlap and their boosts are of roughly the same magnitude.
Overlap is computed between the top ten predictions of each model using simple overlap and rank-biased overlap~\citep[\abr{rbo}]{webber2010rbo}.\footnote{
    If the ordering of predictions amongst the top ten did not matter, the overlap would be acceptable.
    However, as in most \ir{} settings, we \textit{do} care about the order so use a rank-aware metric like \abr{rbo}.
}
As we might expect based on \clipl{} and \teachtextl{} having large boosts on \msvd{}, their predictions also overlap.
This mechanically explains the difference but fails to explain ``why?''

\begin{table*}[h!]
    \centering
    \small
    \begin{tabular}{p{.475\linewidth}p{.475\linewidth}}
        \toprule
        \textbf{MSVD Short Length Captions} & \textbf{MSVD Median Length Captions}\\
        \midrule
        playing panda&a gymnast falls off a balance beam\\
some work&a person is riding a horse\\
a man&a girl is riding a bicycle\\
a baby&two men are pushing an airplane\\
jumping dachsund&the turtle is playing with the cat\\
naah&piano is played by an artist\\
amanplaysaguitar&the girl put stickers on her face\\
a woman&a boy is reading a card\\
camp&a little boy is playing golf\\
plying music&a man is slicing a tomato\\
        \midrule
        \textbf{MSVD Long Length Captions}\\
        \midrule
        \multicolumn{2}{l}{a man holding an open umbrella jumps across a wooden stand in a park and then does a summersault after kicking a wall}\\
\multicolumn{2}{l}{a man in a jail cell motions to a man in another cell who shows the first man his middle finger}\\
\multicolumn{2}{l}{a bowling man picks up a spare in his lane and manages to knock over the one remaining pin in the lane to his right}\\
\multicolumn{2}{l}{a woman is exercising by stepping from right to left and then from left to right while swinging her arms back and forth}\\
\multicolumn{2}{l}{a man wearing a black cape is walking toward a group of people and a man in the group is shooting at him with a pistol}\\
        \bottomrule
    \end{tabular}
    \caption{
        This table shows three sets of \msvd{} captions sampled from: (1) the 100 shortest captions, (2) median length captions, and (3) the 100 longest captions.
        As also observed in \msrvtt{} captions (Table~\ref{tbl:caption_examples_msrvtt}), short captions are general (e.g., ``a man'') compared to the longest captions.
    }
    \label{tbl:caption_examples_msvd_short}
\end{table*}

We test the hypothesis that shorter queries have more positives because they are less specific (i.e., general) and speculate that differences in \clipl{} and \teachtextl{} pre-training could make \clipl{} fare better on general queries.
Intuitively, shorter captions should be less specific and therefore match more videos, so models that handle general captions well should benefit the most.
Table~\ref{tbl:caption_examples_msvd_short} and Appendix Table~\ref{tbl:caption_examples_msrvtt} validate this intuition by showing \msvd{} and \msrvtt{} captions.
The captions are sampled from the shortest 100 captions, median length captions, and longest 100 captions.

\begin{figure*}[t]
    \centering
    \includegraphics[width=\linewidth]{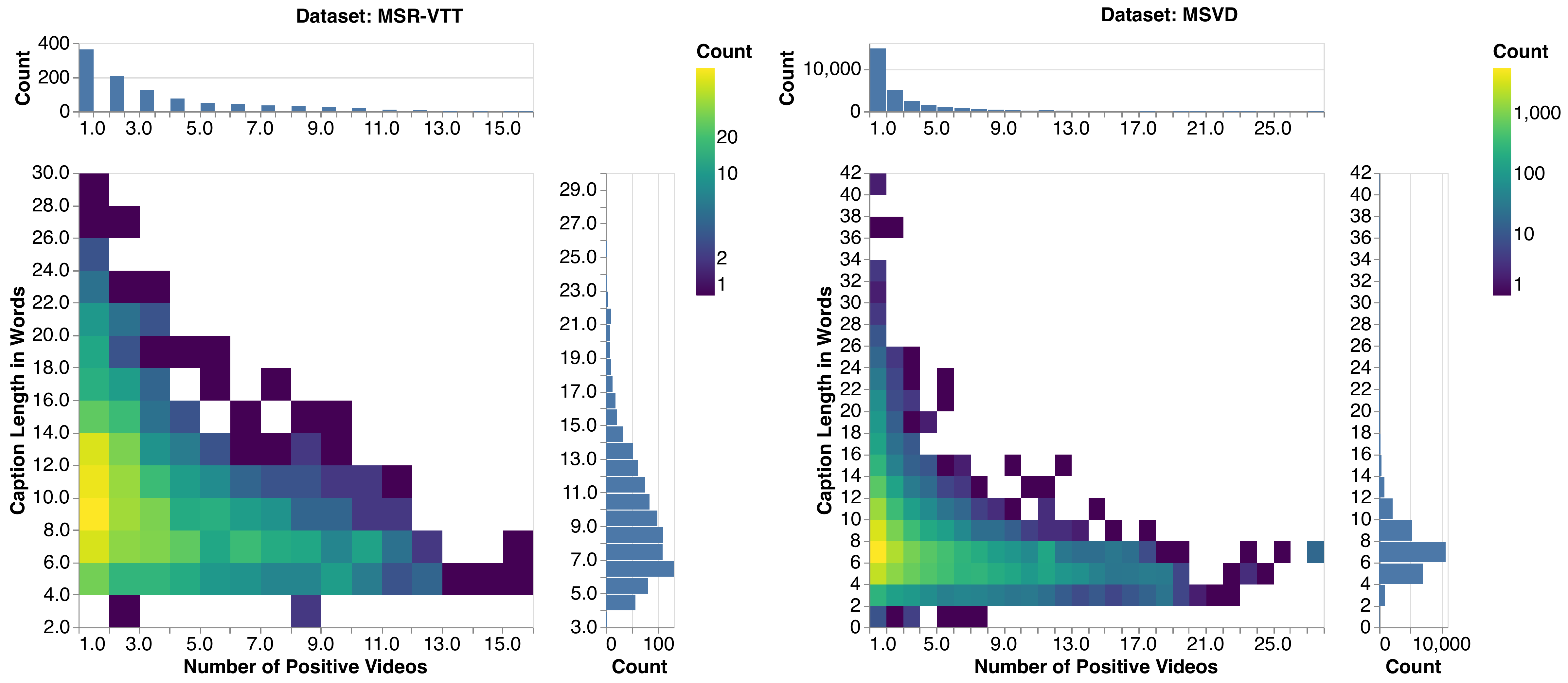}
    \caption{
        This figure shows the relationship between the number of positive videos and the length of captions in words for \msrvtt{} and \msvd{}.
        We show a log-scale density heatmap binned by the number of positive videos and caption length; on the margins, are histograms.
        From this figure, we can infer that: (1) if a caption is long, it is less likely to have many positive videos, and (2) if a caption is short, then the number of positive videos can vary widely.
    }
    \label{fig:positive_v_length}
\end{figure*}

First, we empirically validate that short captions have more positive videos.
Figure~\ref{fig:positive_v_length} shows that longer captions have fewer positive videos while shorter captions have more.
By construction, since we find only positives if a model predicts them, these are where models make gains.

\begin{figure*}[t]
    \centering
    \includegraphics[width=\linewidth]{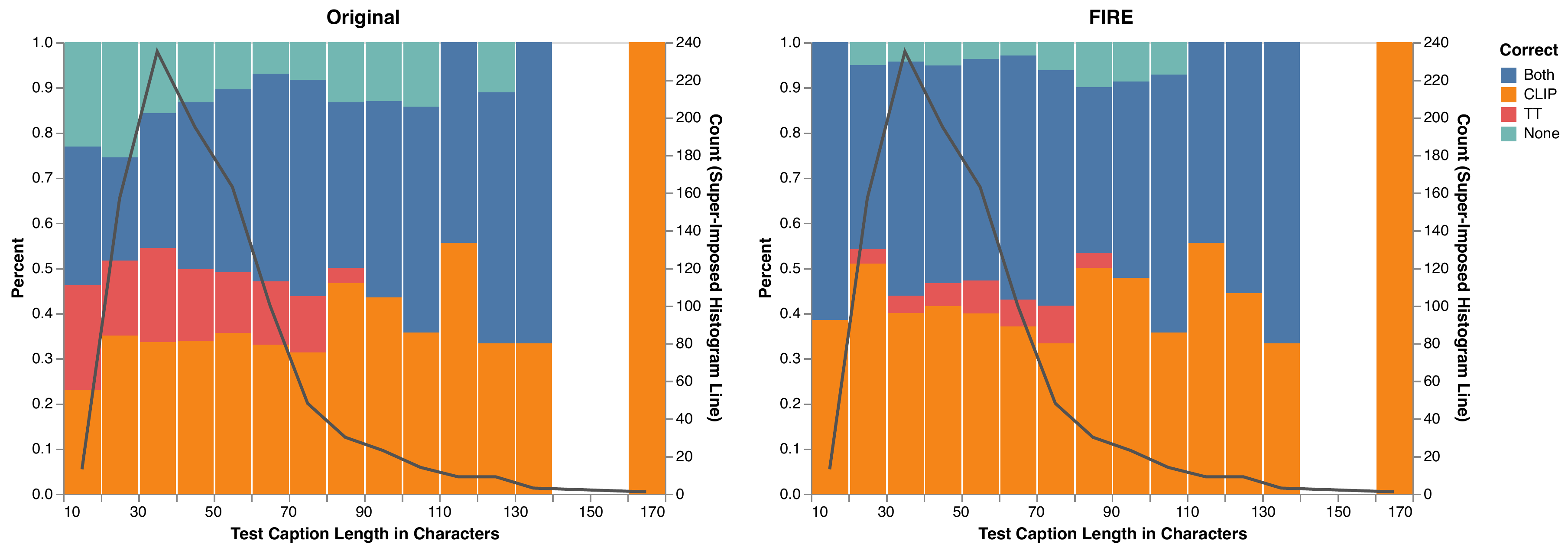}
    \caption{
        On \msrvtt{}, we show relative model effectiveness differences (y-axis and color bars) broken down by test caption length (x-axis); we super-impose the caption length distribution (black line).
        Short captions tend to be more general, so they should match more videos and produce more false negatives.
        The gains for both models and especially \clipl{} occur predominantly on this subset (reduction of ``None'') as we would expect.
    }
    \label{fig:length-perf}
\end{figure*}

Figure~\ref{fig:length-perf} takes the next step and compares model accuracy as a function of caption length.
For each bin of caption lengths (e.g., captions of length zero to twenty characters), we show the proportion of whether both \clipl{} and \teachtextl{} are correct, neither are correct, or only one is correct.
Empirically, we observe that \clipl{} makes the largest gains from accounting for false negatives with \fire{} when queries are short---whether this is due to short queries containing more positives or \clipl{} handling these better is difficult to discern.
Although it is difficult to validate, our best, educated guess at a causal reason for \clipl{} finding more positives in \msrvtt{} is that its image-text backbone, \abr{clip}~\citep{radford2021clip}, was trained with text that contains many general captions.

\subsection{Does System Pooling Generalize?}
\label{sec:pool-gen}

Although system pooling eliminates (implicit) false negatives, it comes with the substantial drawback that every new model must have its predictions annotated---otherwise, the results are potentially biased against the new model due to the possibility of false negatives in novel predictions~\citep{yilmaz2020pooling}.\footnote{
    If a model predicts a video that no prior model does \textit{and} it is a false negative, then the model's effectiveness will be underestimated.
    \citet{yilmaz2020pooling} study this when comparing traditional and deep learning \ir{} systems.
}
System pooling has traditionally been used in synchronized shared tasks where all models are submitted by a deadline and evaluated at the same time, as in the Text REtrieval Conferences (\abr{trec}) in \abr{ir}.\footnote{\url{https://trec.nist.gov}}
However, the trend in machine learning and \nlp{} is for continuously running or even dynamic benchmarks~\citep{kiela2021dynabench}.
Beyond benchmarking, even the development of new models is affected since gains from improved modeling may be understated.
The question then is: how large is this bias, and how fast does it decrease with the number of pooled models?

The magnitude of the bias is affected by two factors: (1) the percent of model predictions that do not exist in pooled annotations and (2) the prevalence of false negatives in this subset.\footnote{
    See Appendix~\ref{sec:pos_dist} for analysis of the number of known positive videos per query.
}
While Table~\ref{tab:overlap} captures prediction overlap between pairs of models, it does not capture the setting where some number of models have annotated predictions, and we wish to test a new model.
Table~\ref{tab:ablation} calculates (1) model scores when using all annotated predictions versus (2) model scores using only annotated predictions from the other two models.
In this small three-model experiment, the bias is unfortunately still significant ($24.4\%$ for C@1) for the best model (\clipl{}).
Thus, the degree to which the \fire{} dataset will mitigate the false negative problem in new model development is dependent on the similarity of new models to current ones.
The generalizability also depends on the number of unknown positives, which we indirectly study by plotting the ranks of positive videos (Appendix~\ref{sec:positive_rank}).

\subsection{Mitigating Annotation Costs by Sampling}
\label{sec:sampling}

A limitation of our method is that until existing annotations include most positives, our method either disadvantages new models or introduces non-trivial annotation costs.
Indeed, the costs of exhaustive annotation in our work are substantial, but exhaustive annotation is also excessive if the goal is only to (robustly) estimate model scores.
Instead, we propose that future work need only annotate the top 10 predictions from $N$ examples in the evaluation data.
%\footnote{
%    Previous work reports scores computed from the top 10 predictions, so we preserve this for comparability.
%}
But how large should $N$ be so that we can be confident that the difference between model scores is statistically significant?
In our next experiment, we use bootstrap sampling to characterize the relationship between $N$ and the effect size corresponding to a statistically significant difference at the 95\% confidence level.

\begin{figure*}[t]
    \centering
    \includegraphics[width=.85\linewidth]{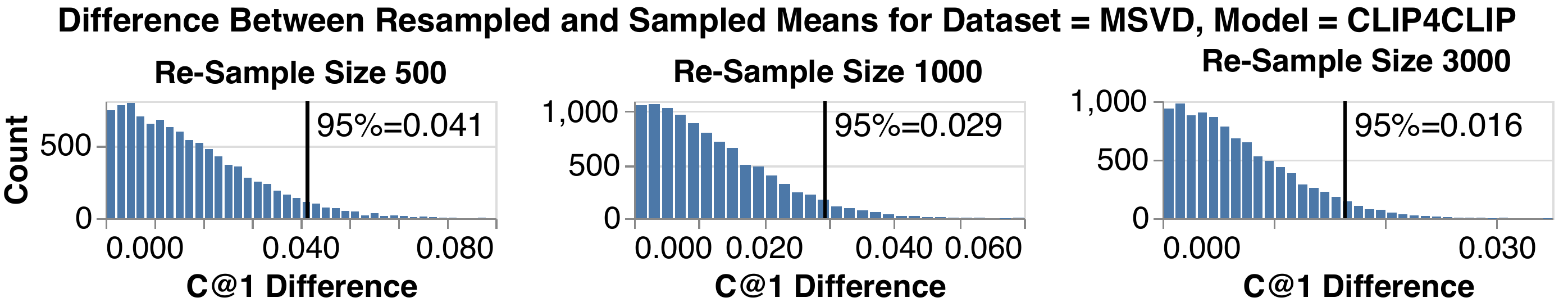}
    \caption{
        The distribution of absolute differences between bootstrap re-sample estimates of \clipl{} C@1 and the true sample mean, by re-sample size.
        This estimates the number of annotations to detect an effect size at 95\% confidence.
        Appendix~\ref{sec:sample-tt} expands on this experiment by showing results for C@5, C@10, and \teachtextl{}.
    }
    \label{fig:bootstrap-clip}
\end{figure*}

In our bootstrap sampling experiment, we treat the 27,763 \msvd{} test examples as a sample from a population.\footnote{
    \msrvtt{} is small. To avoid convergence to the sample mean, bootstrap sizes need to stay low.
}
We characterize the population distribution through bootstrap re-sampling of the original sample.
Specifically, we estimate the absolute difference in model scores that correspond to a statistically significant effect size (i.e., score difference) at the 95\% confidence level.
For each sample size $N\in [500,1000,3000]$, we (1) re-sample $N$ examples from \msvd{} evaluation data, (2) calculate scores on the re-sample, (3) repeat this 10,000 times, (4) average the scores then calculate the absolute value of the difference between the average score and score calculated with the full dataset, and finally (5) plot the distribution and score corresponding to the 95\% percentile.
The experimental results (Figure~\ref{fig:bootstrap-clip}) demonstrate that annotation volumes of 1,000 detect statistically significant differences when C@1 differs by $0.029$.
The results demonstrate that (1) annotating a subset of test examples detects absolute differences of one absolute point, and (2) the number of annotated test predictions varies based on the metric of interest.

\section{Recommendations for Benchmarks}
\label{sec:recs}

Towards improving \mmr{} evaluations, we make recommendations for both current and future benchmarks.
This paper only investigated the effects of false negatives in \msrvtt{} and \msvd{}.
However, it is likely that other similarly constructed benchmarks exhibit the same problem, and testing this is important.
Second, we show that for \msrvtt{} and \msvd{}, certain metrics such as Correct@10 are potentially saturated since improvements above \clipl{}'s $95.7\%$ and $94\%$ are plausibly noisy.
Consequently, since the remaining gains reside in re-ranking the top K, the community should consider retiring these evaluations.
Third, the introduction of multiple positives and use of various K values makes mean average precision attractive since: (1) it factors in preference for correctness at higher ranks and (2) it handles multiple positives.

It is difficult to recommend that model developers exhaustively annotate model predictions.
This suggests a future where query or video set size is a trade-off between annotation load and evaluation quality.
For example, one might choose to trade-off annotation load with statistical power to differentiate between models~\citep{card2020power}.
\abr{trec}-style, annual shared tasks are one model for this~\citep{voorhees2019evolution,church2019eval}; instead of building a monolithic benchmark that becomes overfit over time~\citep{blum2015ladder,andersonCook2019host}, stakeholders develop evaluations that evolve with research objectives.

Looking forward, \mmr{} evaluation would benefit from: (1) a purpose-built benchmark that is grounded in an actual use case so as to be ecologically valid~\cite{devries2020eco} and (2) centralized evaluation by submitting runnable models to shared infrastructure such as Dynabench~\citep{kiela2021dynabench}.
This would improve reproducibility, which was a limiting factor in selecting which model predictions to reproduce in this paper.
This also makes calculating statistical tests easy~\citep{ethayarajh2020utility}, which are often not reported~\citep{dror2018guide,Dodge2019ShowYW}.
%\footnote{
%    In the context of our main experiments, computing statistical significance would be difficult since multiple runs would require additional annotation for each run.
%}
\mmr{} modeling has advanced enough to demand better benchmarks for measuring future progress.

\section{Related Work}
\label{sec:rel}

The paper draws on ideas in multimodal retrieval, information retrieval, and evaluation methodology.

\textbf{Improving Benchmark Quality:}
\citet{wray2021similarity} is directly relevant to our work, and we share their motivation: to study the effects of false negatives in \mmr{} evaluations.
While we share motivation and our works are complementary, our work differs substantially in methods, contributions, and conclusions.
The primary difference is this: our goal is to quantify the difference in absolute metrics that false negatives cause, even if there is no promise the data can be effectively reused in the future; \citet{wray2021similarity} develop automatically runnable proxy measures that improve the reliability of model rankings, but do not precisely quantify the impact of false negatives on existing metrics since automatic labeling is not equivalent to human annotation.
Both these works are valuable: our work conclusively quantifies that false negatives create differences of 25\% absolute points and demonstrate that new measures like those by~\citet{wray2021similarity} are necessary for current benchmarks.
%Their methods are important since they allow for (1) reliable ranking of methods while new benchmarks are developed and (2) reliable estimation of whether models-in-development improve over current methods.

\citet{wang2022multiquery} argue that video captioning datasets used in \mmr{} evaluation are noisy due to low-quality captions but differ by identifying single query tasks as the root problem (as opposed to false negatives) and recommend multi-query evaluation where users make followup refinement queries.
While the multi-query problem is important, we do not agree with the assessment that single-query problems should be abandoned for multi-query problems: for example, users often have a low tolerance for voice assistant errors and abandon their query entirely after an error.
Both problems are important.
Fortunately, the approaches are complementary and should be combined: the multi-query setting still has false negatives, whose effects on measurement can be mitigated with our methods (\S\ref{sec:sampling}).
Just as we use predictions to improve datasets, \citet{beyer2020done} improve ImageNet labels by using predictions to reduce the label space which makes the annotation task easier.

\textbf{Benchmarking:}
Across machine learning, computer vision, and natural language processing~\citep{eger2020workshop,bowman2021fix,rogers2021change} there is a broad effort to critically examine the benchmarks~\citep{schlangen2020targeting}, data~\citep{linzen2020progress,thrush2022dynatask}, evaluation methods~\citep{rodriguez2021leaderboard}, and evaluation paradigms~\citep{rodriguez2021paradigms,kiela2021dynabench} used in research studies.
This effort goes beyond particular methodologies and extends to identifying the values prized by the community~\citep{sculley2018curse,dotan2020value} which are subsequently operationalized in computer vision datasets and benchmarks~\citep{wu2017book,scheuerman2021politics}.
Our work is in line with this broader initiative and critically examines \ttv{} retrieval evaluation methodology.

We examine internal validity~(\S\ref{sec:internal}) and find a broken yardstick~\citep{hernandez2020yardsticks}.
By examining construct validity~(\S\ref{sec:construct}), we also argue that \mmr{} evaluations should prize usefulness to ecologically valid use cases such as real-world text-to-video search~\citep{devries2020eco}.
Lastly, our experimental results suggest we may not be far off from retiring \msrvtt{} and \msvd{} for \mmr{} evaluation, something we should not be afraid to do in general~\citep{boydgraber2020nerds}.
An alternate approach is smaller, periodic evaluations as in \abr{trec}~\citep{smeaton2002trecvideo,Voorhees2000BuildingAQ,mmeaton2009trec}.
%Beyond improving benchmarking in general, our work is also connected to those that improve dataset quality.
Part of the solution is to create purpose-built datasets with clear goals~\citep{gebru2018datasheets,bender2018data} as opposed to continually re-using datasets intended for different uses~\citep{koch2021rrr}.

\textbf{Structurally Similar Tasks:} \mmr{} is not the only evaluation with the implicit false negative problem.
Our critique is applicable to image retrieval benchmarks that use caption-media pairs from image captioning datasets~\cite{lin2014coco,plummer2015flickr} as the only positives~\citep{karpathy2015image,kim2021vilt,singh2022flava}.
%A similar problem is also encountered in question answering where not all sufficiently equivalent documents~\citep{ni2021mitigating} or answers~\citep{chen2020mocha,si2021name,bulian2022answer} are marked correct.
%Although obtaining sufficient annotations is challenging, it is critical to at least identify when this problem threatens benchmark validity.

\section{Conclusion}
\label{sec:conc}

In this work, we show that label errors (false negatives) in \ttv{} retrieval benchmarks invalidate their \textit{internal validity}---the measured metrics do not accurately reflect reality~(\S\ref{sec:state}).
Following this, we critique the applicability of benchmarks to real-world use cases (\textit{construct validity}).
To estimate the impact of false negatives on benchmark metrics, we collect the \fire{} dataset~(\S\ref{sec:improving}) which contains \firesize{} relevance judgements.
Analysis experiments~(\S\ref{sec:exp}) suggest explanations for why \clipl{} scores higher and estimate system pooling generalization.
Based on our findings, we highlight properties that future \mmr{} benchmarks should have and outline approaches to addressing inherent challenges in retrieval evaluation~(\S\ref{sec:recs}).
Finally, we position our work in the broader effort to improve benchmarking by better aligning tasks with the intended use and improving measurement (\S\ref{sec:rel}).

\section{Limitations}

Our work has several notable limitations.
First, our experiments use two representative and commonly used \mmr{} datasets (\msrvtt{} and \msvd{}).
While we expect that our results will generalize, it is still possible that these results do not generalize.
For example, both datasets are based on YouTube videos and annotator-written captions: perhaps videos and captions from alternate sources differ by too much.
Similarly, our experiments use three well-known models, so while we expect our results to generalize to similar models, future models may differ substantially in ways that cause the empirical results not to hold.
This said, system pooling has long been used in \abr{trec}~\citep{voorhees2005trecbook}, so we expect this to work for future models as well.

Beyond limitations in generalizability, the in-principle critiques in our work apply only to benchmarks where implicit false positives are likely to be prevalent; it does not apply to benchmarks in general.
From the methods perspective, while our computational experiments are coded to be easily reproduced, the scale of our annotations is difficult to reproduce (hence limited reproducibility in this sense), but we do study sampling-based alternatives to mitigate this limitation.

\section{Ethics}

This section discusses potential ethical issues related to our dataset-centric work.
First, we discuss data-related ethics.
The \fire{} dataset is built on \msrvtt{} and \msvd{}.
We distribute the minimal amount of data related to these datasets necessary to reproduce our experiments: triplets of caption identifiers, video identifiers, and annotated labels.
Section~\ref{sec:improving} and Appendix~\ref{apx:annotation} describe how the data was collected.
All annotators were compensated, and the data collection was reviewed before starting.
Potential risks due to the use of our dataset are limited by the additional labels we provide for an existing dataset.
We thoroughly discuss the risks associated with negatively influencing benchmark reliability (i.e., prediction overlap with future models), and these risks are mitigated by our recommendation that more appropriate datasets be developed.

Our work does not directly have negative societal impacts, but it is feasible that the improved model scores we report could be used to misrepresent the capability of retrieval systems.
For example, while we only claim that a model achieves a particular measure of effectiveness on a particular benchmark, the media often inflates the importance of these metrics~\citep{cuthbertson-18}.
In our work, we intentionally do not connect these higher metrics to more general capability and emphasize the importance of establishing construct validity.

\section*{Acknowledgements}

We thank Yookoon Park, Prahal Arora, and Bernie Huang for providing code and data that were helpful to kickstart this project.
We thank Daniel Haziza for infrastructure support in hosting live demos.
We thank Nathan Tokala for their support with annotation infrastructure.
For insightful discussion and ideas, we thank Simran Motwani, Patrick Lewis, Thomas Hayes, Joe Barrow, Xilun Chen, Chenglei Si, Ronghang Hu, Max Bain, and Jacob Kahn.
For feedback on prior versions of this paper, we thank Rich James, Florian Metze, Peter Rankel, Weijia Xu, Yoo Yeon Sung, Yuandong Tian, John P. Lalor, and Kirmani Ahmed.

\bibliography{bib/journal-full,bib/pedro}
\bibliographystyle{style/acl_natbib}

\clearpage
\begin{appendix}
  \section{Model Prediction Comparison}

\begin{figure*}
    \centering
    \tikz\node[draw=black!40!lightblue,inner sep=1pt,line width=0.3mm,rounded corners=0.1cm]{
    \includegraphics[width=.6\linewidth]{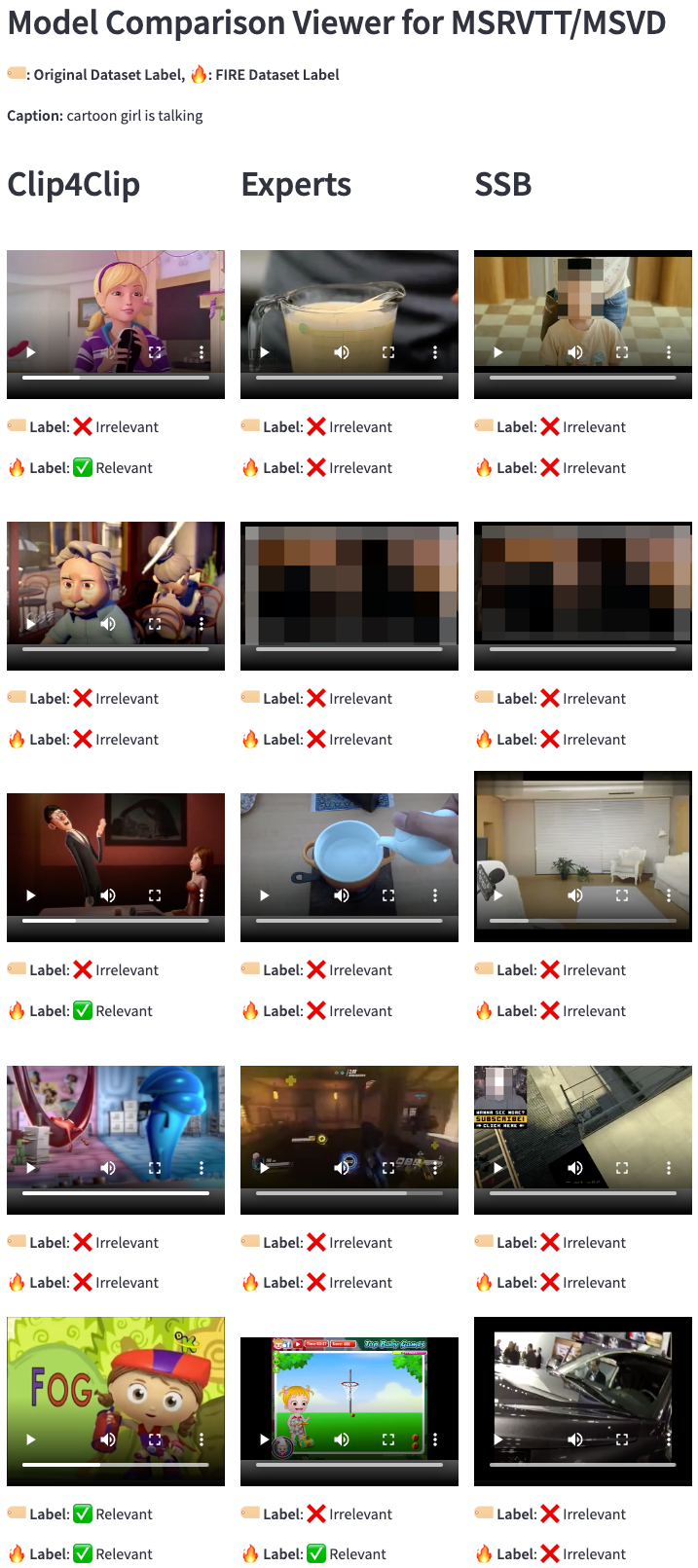}
    };
    \caption{
        The web application shows the ranked predictions of three models: \clipl{}, \teachtextl{}, and \ssb{}.
        Qualitatively, \clipl{} predictions better match the query by showing only cartoon videos.
        This is reflected quantitatively when \fire{} labels are incorporated.
        Lastly, the ranked predictions also show some of the overlap that \teachtextl{} and \ssb{} shared.
    }
    \label{fig:streamlit-compare}
\end{figure*}

As part of this paper, we develop several web apps to make exploring the data more accessible.
For example, Figure~\ref{fig:streamlit-compare} compares the predictions of three models along with the labels in the original \msrvtt{} dataset compared to augmenting them with \fire{}'s labels.
The source code repository provides instructions to run these web app demos.

\section{Annotation Interfaces}
\label{apx:annotation}

\begin{figure*}
    \centering
    \tikz\node[draw=black!40!lightblue,inner sep=1pt,line width=0.3mm,rounded corners=0.1cm]{
    \includegraphics[width=\linewidth]{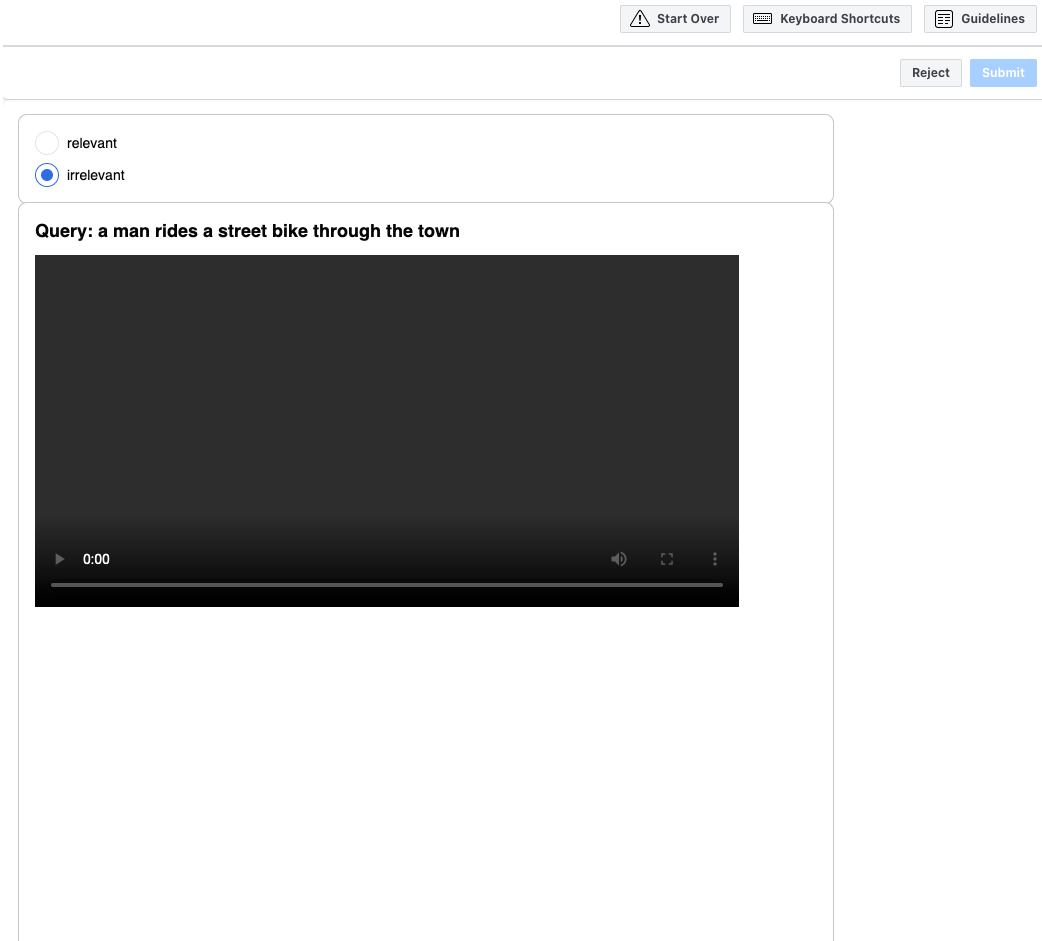}
    };
    \caption{
        To annotate the \fire{} dataset, raters used this annotation interface.
        The interface shows the candidate query (caption) and video; raters are trained to select ``relevant'' or ``irrelevant'' based on whether every component of the query matches the video.
    }
    \label{fig:ui}
\end{figure*}

The \fire{} dataset~(\S\ref{sec:improving}) was collected using the annotation interface in Figure~\ref{fig:ui}.

In addition to the previously described annotation instruction~(\S\ref{sec:dataset}), raters were also instructed on how to handle sensitive categories.
The raters were instructed to accept the caption as accurate unless they had compelling, concrete reasons to believe otherwise (e.g., a little baby should be not considered old, and octogenarians with white hair and wrinkled skin should not be considered young); raters should not attempt to make more fine-grained distinctions.
    In particular, they were instructed not to make any assumptions about gender and accept the gender described by the caption. 

% \section{Caption Statistics}
% \begin{figure}
%     \centering
%     \includegraphics[width=\linewidth]{caption_char_length}
%     \caption{
%         The length distribution (in characters) of captions in \msrvtt{} and \msvd{}.
%         \prtodoi{Add MSVD statistics to the plot}
%     }
%     \label{fig:caption_char_length}
% \end{figure}

\section{FIRE Data Quality}
\label{apx:quality}

\begin{figure*}
    \centering
    \includegraphics[width=\linewidth]{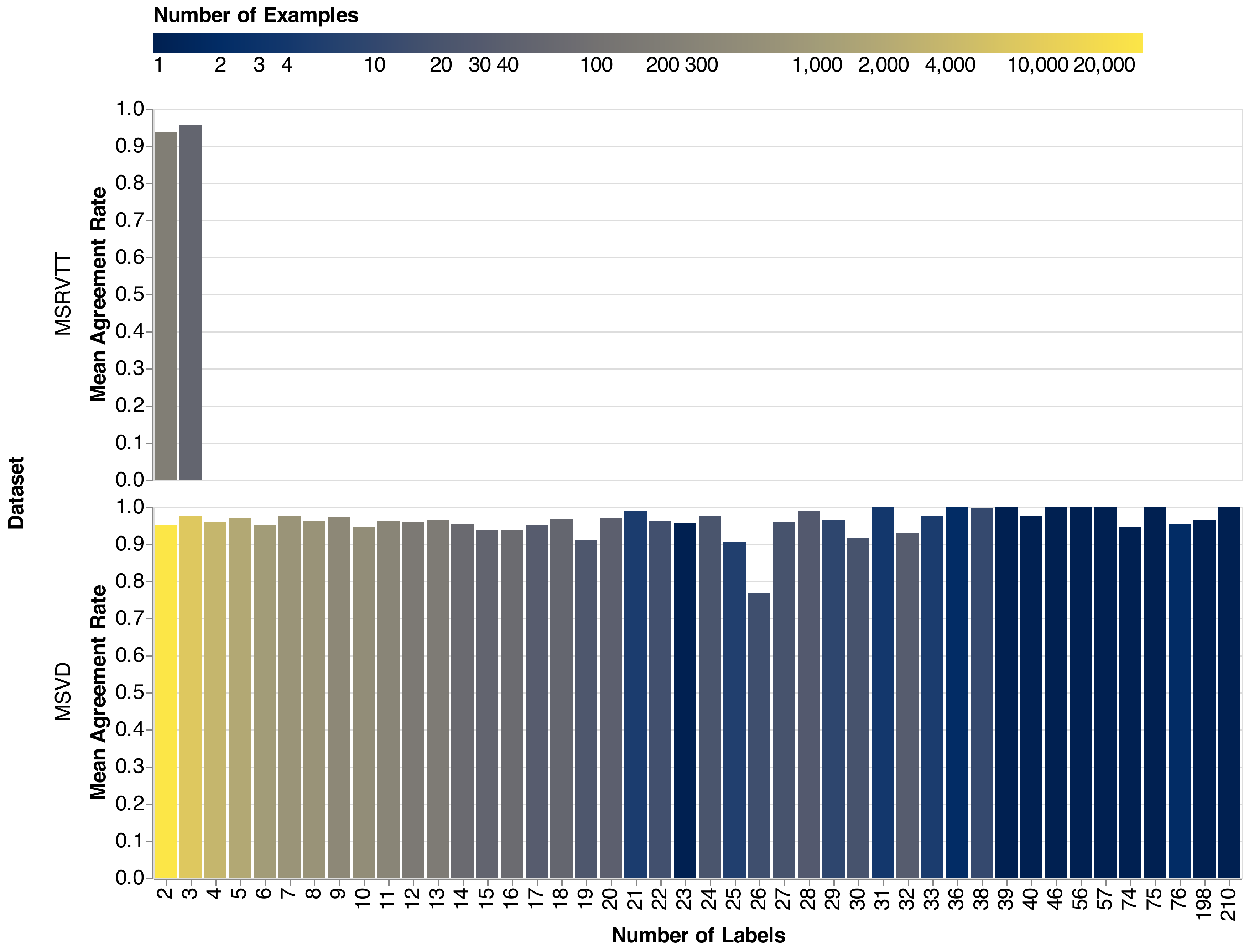}
    \caption{
        The agreement rate of annotators is broken down by the number of labels.
        For example, about 10,000 \msvd{} examples (text-video pairs) were annotated twice; of those, the two labels agreed on about 95\% of examples.
        As we did with the \msrvtt{} collection, our intent for the data collection was to de-duplicate text-video pairs and only annotate about 10\% of the data multiple times to estimate reliability.
        However, we accidentally omitted this step for the \msvd{} collection which resulted in some examples being annotated many times.
        Fortunately, this provides an unplanned opportunity to further validate inter-annotator agreement.
    }
    \label{fig:agree_rate}
\end{figure*}

This section provides additional evidence to validate the quality of the \fire{} dataset.
Specifically, Figure~\ref{fig:agree_rate} complements the agreement metrics computed in \S\ref{sec:quality} and Table~\ref{tbl:stats} by un-aggregating agreement rates.

\section{Shorter Captions, Their Generality, and Correlation to Model Behavior}
\label{sec:lengths}

Experiments in \S\ref{sec:boosts} establish that shorter captions have more positives and longer captions have fewer.
We intuitively explain this by stating that shorter captions by nature are less specific, so will, in principle, match more videos.
For example, one of the shortest captions in \msvd{} is ``a man'' (Table~\ref{tbl:caption_examples_msvd_short}) which is less specific than one of the longest captions like ``a man holding an open umbrella jumps across a wooden stand in a park and then does a summersault after kicking a wall.''
Inspecting these captions also validates our construct validity critique (\S\ref{sec:construct}): they do seem like search queries.

\begin{table*}[t]
    \centering
    \small
    \begin{tabular}{p{.95\linewidth}}
        \toprule
        \textbf{MSR-VTT Short Length Captions}\\
        \midrule
        a man playing video games\\
anchor talking about a shows\\
a woman is stirring food\\
sports are being played\\
a woman holding a ribbon\\
a diver goes underwater\\
baseball player hits ball\\
cartoon show for kids\\
two women are embracing\\
advertisement of seat basket\\
        \midrule
        \textbf{MSR-VTT Median Length Captions}\\
        \midrule
        a man runs into the crowd when trying to catch a basketball\\
in a music video a man is laying with women while singing\\
some people video conferencing as they watch a movie\\
a boy is trying out for a part on the voice kids\\
basketball players making a shot in the last seven seconds\\
views of two persons working on the super computer with the head phones on\\
a character is jumping and floating in the air in a video game\\
two people playing basketball and the one with a hat makes every shot\\
batman is beating up bane in a scene from a batman movie\\
a girl being surprised with a stuffed animal by male friend\\
        \midrule
        \textbf{MSR-VTT Long Length Captions}\\
        \midrule
        a man and a woman are sitting in front of a television and addressing and audience\\
        a woman stirs up some soup sprinkles a spice in and drops a shot of liquid into it\\
        a man is filming as he and a woman watch the news where it shows an area filled with smoke\\
        flight is shaken and the pilots trying to land the flight while they opened the air\\
        the chef adds fish sauce and fish paste to a large stainless steel cooking pot\\
        a girl wearing a dress stands to the side of the screen while lyrics to a song playing in the background appear on the other side\\
        the man is giving an informational speech to a group of people about telling someone something\\
        a girl in blue color dress wearing siting speaking and television screen with black shirt man beside still image displaying on screen\\
        a man plays a video game where the player has a first person perspective and shoots other characters\\
        a man playing a video game character that is carrying a sword and killing animals with it\\
        \bottomrule
    \end{tabular}
    \caption{
        This table shows three sets of captions from \msrvtt{} sampled from: (1) the 100 shortest captions, (2) the median length captions, and (3) the 100 longest captions.
        As we argued by intuition (\S\ref{sec:boosts}), inspecting these samples validates that the shorter captions are more general (e.g., ``sports are being played'') and longer captions are very specific (e.g., ``a woman stirs up some soup sprinkles a spice in and drops a shot of liquid into it'').
    }
    \label{tbl:caption_examples_msrvtt}
\end{table*}

\begin{figure*}[t]
    \centering
    \includegraphics[width=\linewidth]{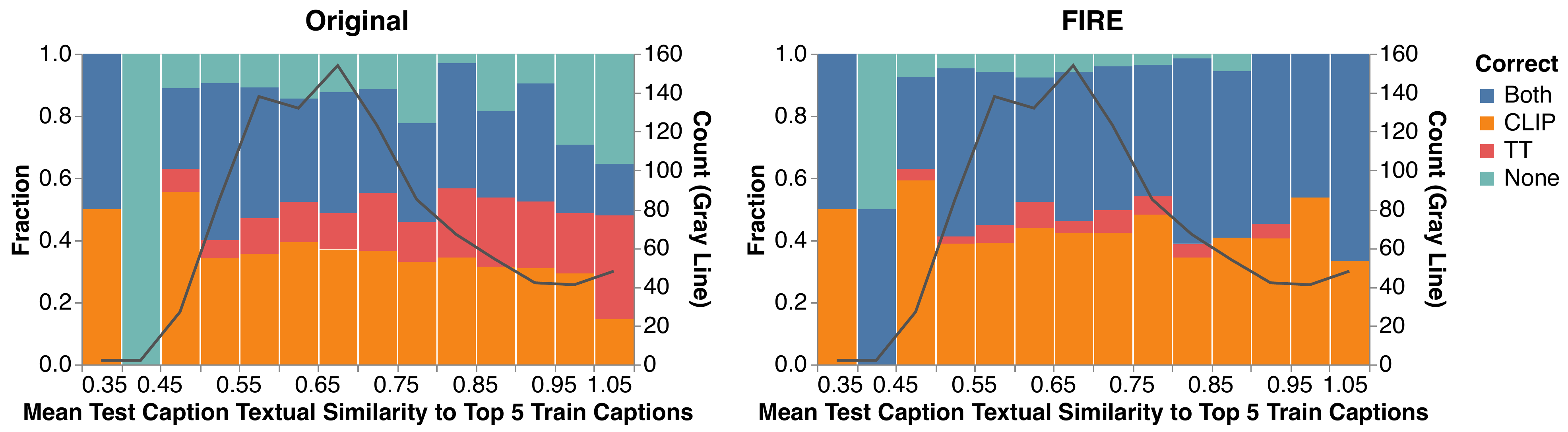}
    \caption{
        Why are the \msrvtt{} score differences between \clipl{} and \teachtextl{} when using \fire{} large?
        We test the hypothesis that \teachtextl{} (comparatively) overfits textual training data.
        We compute the textual similarity of each test caption to each train caption with a 5-gram character model; for each test caption, we calculate the mean similarity of the ten most similar captions.
        The plot shows whether both models score a point on Correct@5, binned by train-test similarity (the overall histogram is shown as the super-imposed line) when using original versus \fire{} annotations.
        On the original annotations, \clipl{} fares much better compared to \teachtextl{} when similarity is not nearly 1.0 (i.e., not overfitting).
    }
    \label{fig:train-test}
\end{figure*}

In previous experiments (\S\ref{sec:boosts}), we discussed how caption length is related to which models gain higher boosts.
This section breaks down which models gain the most on \msrvtt{} by train-test overlap.
We take inspiration from question answering and language modeling, where unintentional textual overlap between train and test sets degrades evaluation and model quality~\citep{lewis2020overlap,borgeaud2021retro,lee2022dedup}.
Our objective is to measure the degree to which test captions in \msrvtt{} are present in the training captions---be it word-for-word or approximate.
To measure this, we use Scikit-Learn~\citep{scikit-learn} to fit a 5-gram character \abr{tf}-\abr{idf} encoder to the test captions and compute the cosine similarity of each test caption to each train caption.
For each test caption, we compute the mean similarity of the top train ten captions and combine this information with Correct@5 scores~(Figure~\ref{fig:train-test}).
The results suggest that \teachtextl{} overfits the train set, which may explain its comparatively better scores on the original positives---were it not overfit, train-test overlap should not matter.
%That \clipl{} has better scores on \fire{} positives suggests that it generalizes better to non-train pairs.
%In short, it seems that \clipl{} text encodings generalize better, and this is one source of its improved scores.

Lastly, these factors are not unrelated.
Since shorter captions tend to be less specific, these are also the captions that we would expect are more prevalent in the training set, whether in exact form or approximate (e.g., the phrase ``a man'' is likely in the train set).
To test whether these factors are related, we compute the Kendall Tau correlation and Spearman Rank correlation between the train-test textual similarity score and caption length (in both words and characters).
As we expect, there is a non-trivial negative correlation between caption length and similarity score (Table~\ref{tbl:correlation}): the lower the caption length, the higher the train-test overlap score.

\begin{table*}[t]
    \centering
    \small
    \begin{tabular}{l l r}
        \toprule
        \textbf{Length} & \textbf{Spearman} & \textbf{Kendall}\\
        \midrule
        Word & $-0.419$ & $-0.296$\\
        Character & $-0.479$ & $0.334$\\
        \bottomrule
    \end{tabular}
    \caption{
        This table shows the Spearman and Kendall rank correlations between the train-test textual similarity score used to measure train-test overlap (Figure~\ref{fig:train-test}) and the length of captions in both words and characters.
        The results support our hypothesis that caption length and train-test overlap are correlated.
    }
    \label{tbl:correlation}
\end{table*}

\section{Can Annotation Costs be Mitigated Through Sampling?}
\label{sec:sample-tt}
\begin{figure*}[t]
    \centering
    \includegraphics[width=\linewidth]{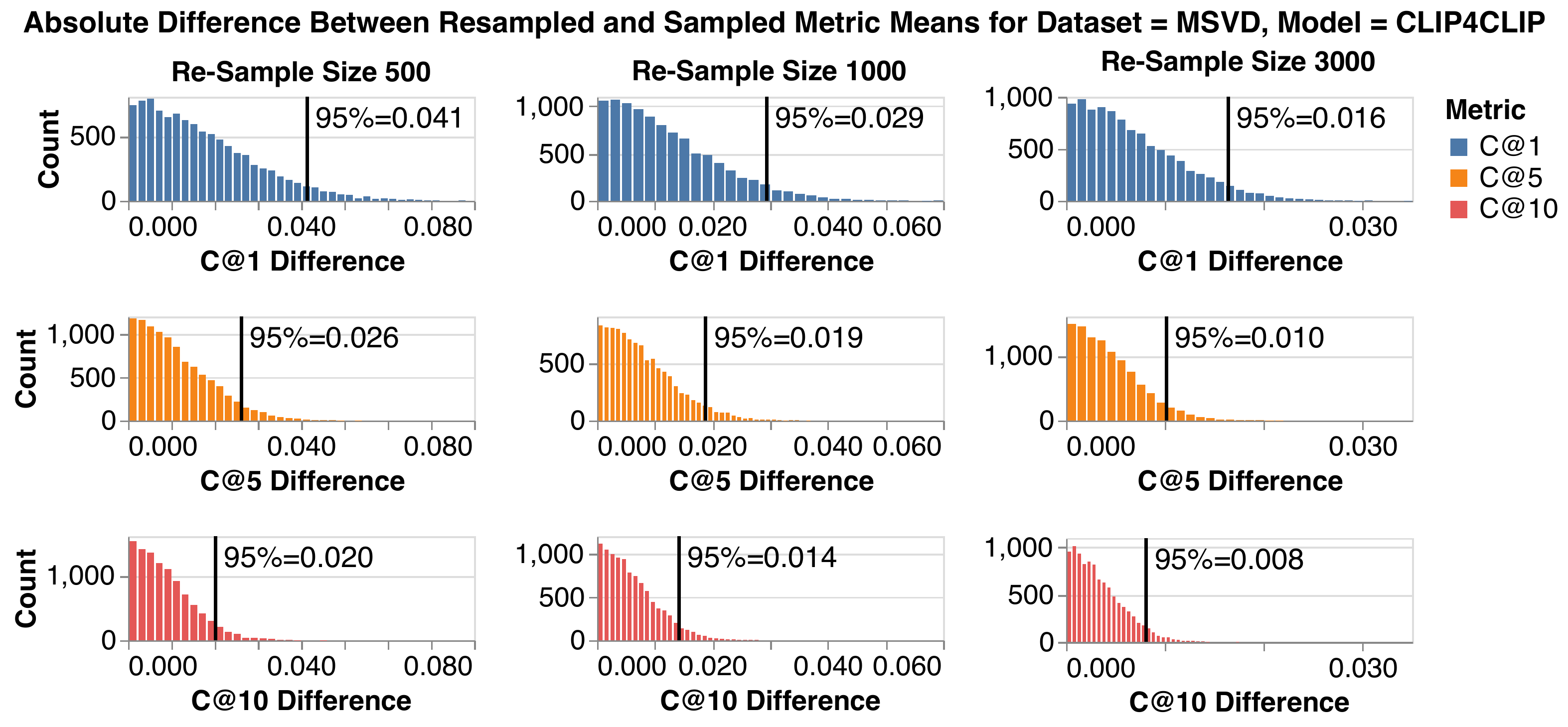}
    \caption{
        This figure replicates the C@1 results from Figure~\ref{fig:bootstrap-clip} but adds results for C@5 and C@10.
        The additional results are consistent in showing that differences of about 1 point are already detectible with 1,000 annotations.
    }
    \label{fig:bootstrap-clip-full}
\end{figure*}

\begin{figure*}[t]
    \centering
    \includegraphics[width=\linewidth]{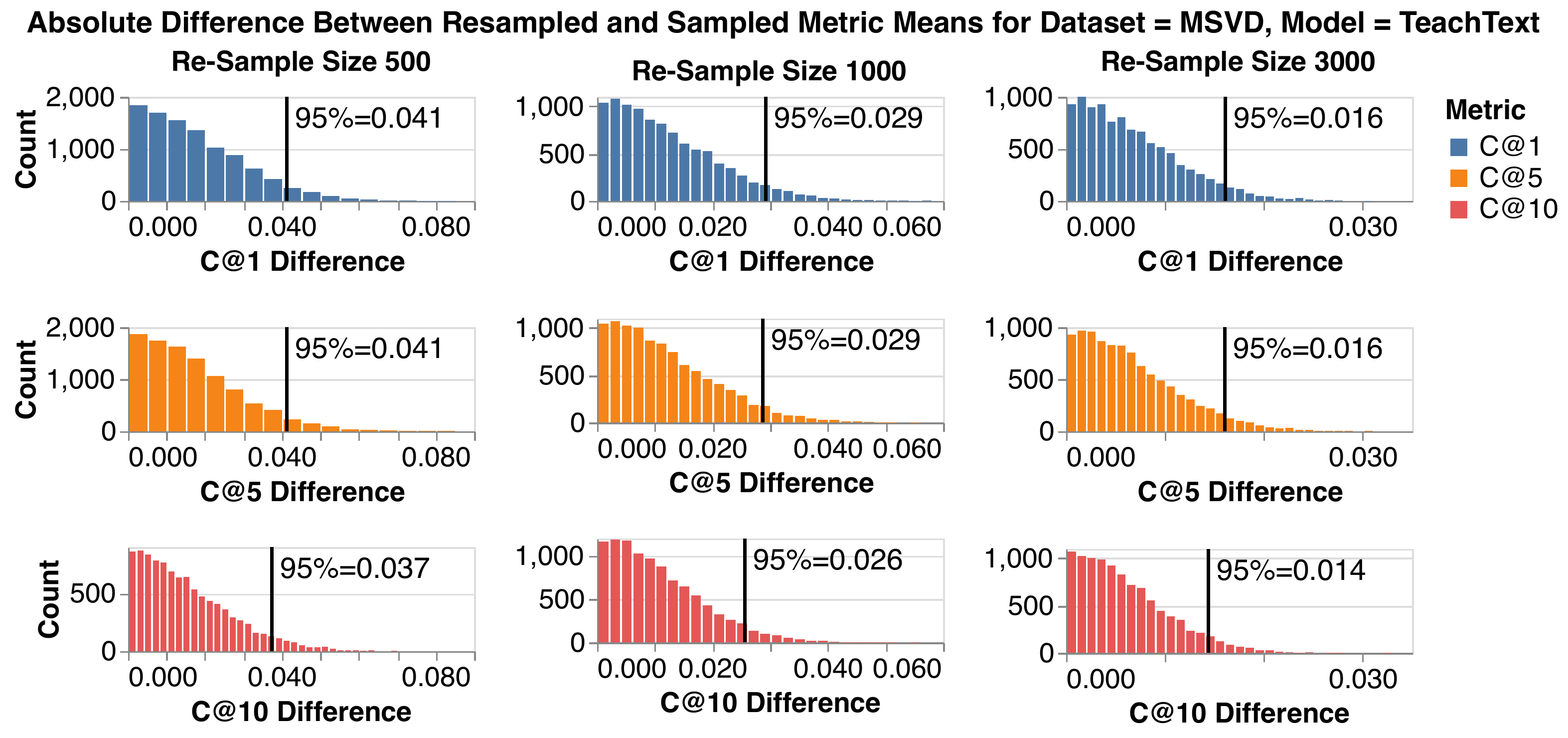}
    \caption{
        Similar to Figure~\ref{fig:bootstrap-clip}, this figure shows the distribution of absolute differences between bootstrap re-sample estimates of \teachtextl{} C@1, C@5, and C@10 scores and their true sample mean (i.e., scores on the full test set).
        Compared to \clipl{}, statistically significant differences are marginally harder to detect.
    }
    \label{fig:bootstrap-tt}
\end{figure*}

In our experiments, we use bootstrap sampling to estimate the number of example annotations needed to detect given effect sizes at the 95\% confidence level~(\S\ref{sec:sampling}).
Figure~\ref{fig:bootstrap-clip} reports these results for \clipl{} since it was the best model; in practice, it represents the type of model we would test after models like \teachtextl{}.
Figure~\ref{fig:bootstrap-clip-full} extends the results from C@1 to C@5 and C@10.
Figure~\ref{fig:bootstrap-tt} replicates these results, but using the  \teachtextl{} model.

\section{Number of Positive Videos per Text Query}
\label{sec:pos_dist}

\begin{figure*}[t]
    \centering
    \includegraphics[width=\linewidth]{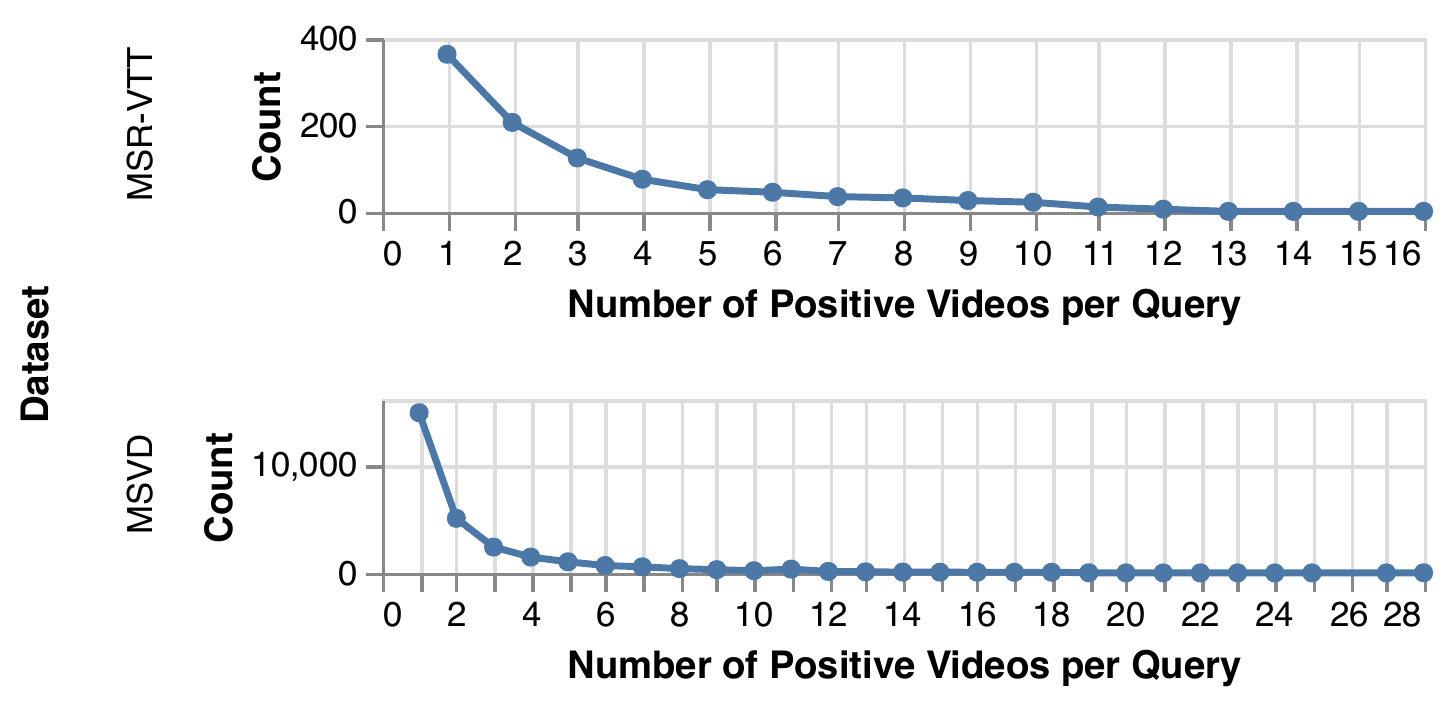}
    \caption{
        For \msrvtt{} and \msvd{}, we plot the number of positive videos per test set query.
        While many queries across both datasets have only one known positive, many others have more than that.
    }
    \label{fig:n_positive}
\end{figure*}

The generalization of the \fire{} dataset to newer models is reliant on two factors: (1) the number of positive videos per query and (2) whether the models we studied in this work predict all the true positives.
Estimating the number of true positives per query without exhaustive annotation is difficult at best.
However, we can at least characterize how many positives there are when including \fire{} annotations.
Figure~\ref{fig:n_positive} shows a histogram of the number of positive videos per query across \msrvtt{} and \msvd{}.
For example, about 350 \msrvtt{} queries have only one known positive, which implies that the other 650 have more than one known positive.
Unfortunately, even estimating the upper bound would require annotating all the videos for each query in a representative sample (e.g., for a sample of \msrvtt{} 200 queries, exhaustive annotation would include $\text{200}*\text{1,000}=\text{200,000}$ query-video annotations).

\section{Rank of Positive Videos}
\label{sec:positive_rank}

\begin{figure*}[t]
    \centering
    \includegraphics[width=\linewidth]{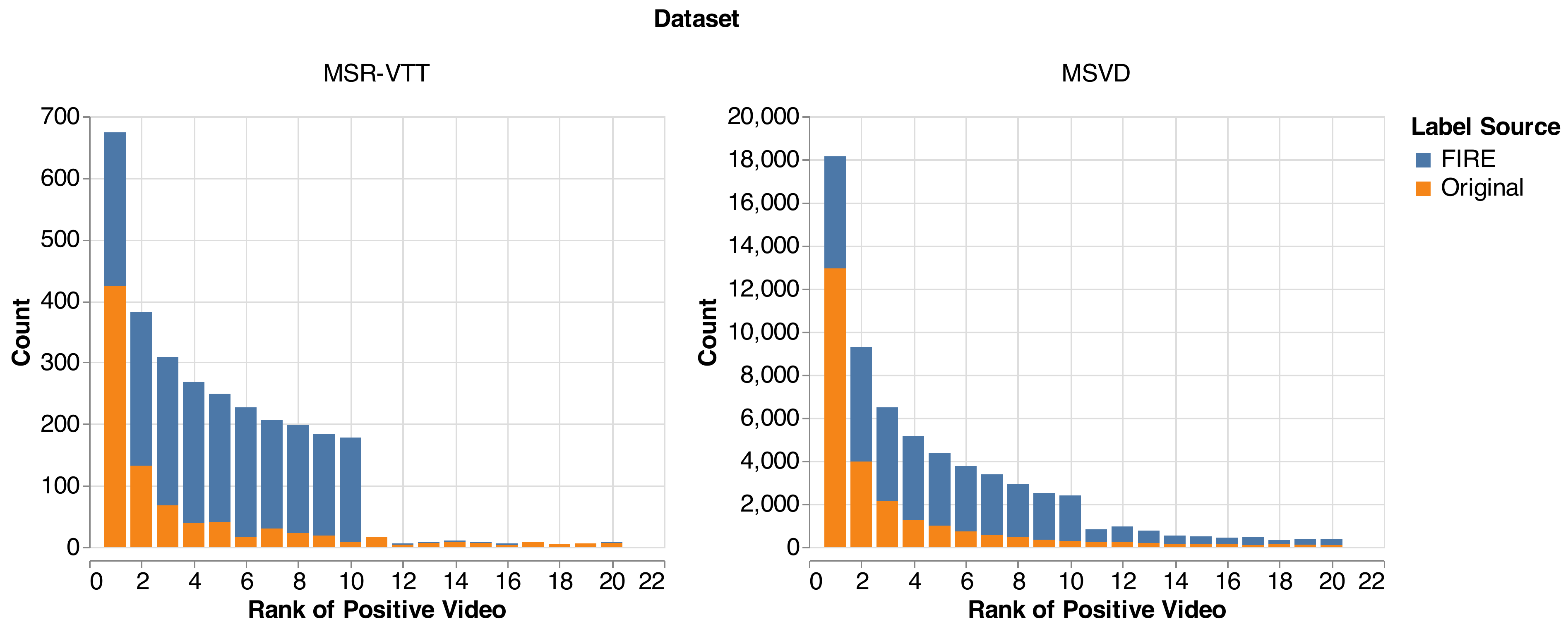}
    \caption{
        The figure plots the rank of positive video predictions from \clipl{} versus their count.
        The plot displays \msrvtt{} and \msvd{} separately, and it breaks down the source of each positive (from the original dataset versus from \fire{}).
        While the distribution suggests most positives are found within the top 10, the long tail suggests that there are still unknown positives.
    }
    \label{fig:positive_rank}
\end{figure*}

While we follow prior work in measuring models based on metrics computed from their top ten predictions, this still leaves open the question: ignoring prior work, is ten predictions the right choice?
If ten is the correct choice, then we should see a clear trend that positives are primarily distributed below ten.
Figure~\ref{fig:positive_rank} plots the rank of positive videos in \clipl{} predictions (i.e., 1 is top-ranked) versus their count.
As expected, the number of positives drops dramatically before rank 10 (especially for \msvd{}), although not to zero; the ranks of the original positives suggest there is a long tail of undiscovered positives.
Note that the steep dropoff at 10 is due to annotating only the top 10; positives beyond this are either from the original dataset or predicted by other models.
From this, we conclude that although most positives have likely been collected, there likely remain more past rank 10, especially in \msrvtt{}.

\section{Computational Resources}

This paper was developed using two types of computational resources.
To rerun \ttv{} retrieval models, we trained and evaluated on a single \abr{aws} p4d compute node which has 96 v\abr{cpu}s, 1152GB of \abr{ram}, and eight Nvidia A100 \abr{gpu}s.\footnote{\url{https://aws.amazon.com/ec2/instance-types/p4/}}
All other experiments were run locally on a 16 inch, 2019 Macbook Pro with a 2.4GHz 8-core Intel Core i9 \abr{cpu} and 32GB of \abr{ram}.

\end{appendix}
\end{document}